\documentclass{INTERSPEECH2023}

\usepackage{multirow}
\usepackage{amsmath}
\usepackage{array}
\usepackage{multirow}
\usepackage{cleveref}
\usepackage{makecell}

\interspeechcameraready

\title{Navigating the Minefield of MT Beam Search in Cascaded Streaming Speech Translation}
\name{Rastislav Rabatin{$\dagger$}, Frank Seide, Ernie Chang
 \thanks{*Equal contribution.  {$\dagger$}Corresponding author.}}
\address{
Meta, USA
\email{\{rasto2211,seide,erniecyc\}@meta.com}
}
\begin{document}

\maketitle
 
\begin{abstract}

We adapt the well-known beam-search algorithm for machine translation to operate in a cascaded real-time speech translation system.
This proved to be more complex than initially anticipated, due to four key challenges:
(1) real-time processing of intermediate and final transcriptions with incomplete words from ASR,
(2) emitting intermediate and final translations with minimal user perceived latency,
(3) handling beam search hypotheses that have unequal length and different model state,
and (4) handling sentence boundaries.
Previous work in the field of simultaneous machine translation only implemented greedy decoding.
We present a beam-search realization that handles all of the above, providing guidance through the minefield of challenges.
Our approach increases the BLEU score by 1 point compared to greedy search, reduces the CPU time by up to 40\% and character flicker rate by 20+\% compared to a baseline heuristic that just retranslates input repeatedly.


\end{abstract}

\section{Introduction}

Streaming decoding is an essential part of cascaded real-time speech translation systems. These systems consist of (1) automatic speech recognition (ASR) to transcribe the source speech, followed by (2) a text-to-text machine translation (MT) system. To achieve minimal latency, both must be {\em streaming models} that are architected and trained to emit results as early as possible.
Specifically, the MT must be a {\em Simultaneous MT} model that generates translations early without having to see an entire sentence or paragraph.

Several Simultaneous MT architectures have been proposed, including Monotonic Multihead Attention (MMA) \cite{mma}, Monotonic Chunkwise Attention (MoChA) \cite{mocha}, Monotonic with Infinite Lookback attention (MILK) \cite{milk}, and EMMA \cite{emma}.
While these papers focused on the modeling aspect of the problem, they only implemented greedy decoding.



We introduce a realization of beam-search for use in a real-time pipeline that transcribes and translates audio in real-time, where translations are either shown on a display in a real-time or played
to a loudspeaker using text to speech (TTS).

The fundamental difference from non-streaming beam search is that we need to provide intermediate translations with minimal user perceived latency.
Additionally, we need to process intermediate and final
transcriptions with incomplete words in real-time, and correctly propagate them through the system.
Furthermore, the streaming nature of the problem brings additional challenges
that come with handling multiple beam search hypotheses, and sentence boundaries.
We tackle all of these problems one-by-one, and provide solutions to them in \Cref{streaming-mt-beam-search}.



We compared our approach with a heuristic ``fake streaming'' approach that simulates streaming by repeatedly translating sentence prefixes with a non-streaming model (\Cref{fake-streaming-baseline}).
This approach serves as a good baseline for display output but is not suitable for TTS output because it has high user perceived latency, and is computationally inefficient.
Our cascaded speech translation system needs to run on a wearable
device with a limited computational power which makes the computational efficiency crucial.
We discuss the advantages and disadvantages of different
translation outputs, and explain how our approach addresses them.

The presented algorithm achieves the following:
\begin{itemize}
\item
  Increase in BLEU score by 1 point compared to greedy for beam size 3.
\item
  Up to almost 40\% CPU time reduction compared to repeated retranslations baseline (\Cref{fake-streaming-baseline}).
\item
  20+\% reduction of character flicker compared to system with repeated retranslations.
\end{itemize}


\hypertarget{live-translation-architecture}{%
\section{Background}\label{live-translation-architecture}}

A cascaded streaming speech translation system is composed of several components: real-time ASR, simultaneous MT,
and TTS or display. ASR
receives a live audio stream from microphone(s), and generates a stream of
transcription events which may be \textit{intermediate} or \textit{final}. MT processes the stream of transcription events, and
generates a corresponding stream of translation events (also \textit{intermediate} or \textit{final}) which are either sent
to the display, or to TTS to generate an audio stream to play
via a loudspeaker.

\hypertarget{terminology---intermediates-and-finals}{%
\subsection{Terminology: Intermediates and Finals}\label{terminology---intermediates-and-finals}}

To understand our algorithm and its complexity, we need to introduce the core concept of {\em intermediate} and {\em final}
events that flow through the system.
ASR is feeding MT two types of \textbf{events},
\textbf{intermediates} and \textbf{finals.} \emph{Intermediate} events
denote an ASR result that is not stable---transcription can change.
ASR has not yet made the final
determination, and every intermediate {\em will} be followed by a subsequent event that can be either another intermediate or a final, to substitute words from the starting point of the last intermediate.

\emph{Final} events denote ASR results
that won't change anymore, and they advance the ``starting point of the last intermediate'' to after the final event.
The concatenation of all final events is the final
transcription of the entire utterance. One observes that frequently, events consist of a single newly-recognized word, or (with word-piece based ASR) sometimes even {\em only a prefix of a word}---consider a
plural noun reported as two consecutive final events: its singular base
form followed by an ``s''.


A key goal of our streaming beam search for MT is that it gracefully handles intermediate events and word pieces, by both generating them, but also correctly handling intermediate ASR events.\footnote{How ASR generates intermediates vs.~finals is a well-understood problem that is not subject of this paper.}
An MT intermediate is just like ASR---a
preview as to how one or more words would be translated; but future
additional input may change or refine the word choices.

When the translation target is a display, displaying intermediate translations early is useful, as they provide a perception of responsiveness. If after receiving more context, the MT model decides that an intermediate was incorrect, the display is rewritten, which some users describe as ``intelligent'' behavior.

When a display is not available and the translation target is audio/TTS, however, ``rewriting'' is not possible. Intermediates are not useful, as we must hold
back TTS until translated words have been finalized (are stable). Hence, in this scenario, the
objective for the MT is to finalize and emit translated words as
soon as possible---this is called \emph{streaming MT}.

\hypertarget{fake-streaming-baseline}{%
\subsection{Baseline: Fake Streaming by Repeated Retranslation}\label{fake-streaming-baseline}}

One can approximate simultaneous MT with a non-streaming model
with a traditional encoder-decoder MT model trained to take a whole sentence as its input, where intermediate translations are generated by repeatedly translating the growing prefix of a sentence as it is being spoken.
The updated translations can be updated on the display, sometimes called ``screen rewriting.''

One disadvantage is the substantial flicker because
the translation will evolve as more context becomes available, often replacing big chunks of the text.
The simple solution---simply wait until the end of the sentence--however leads to poor (slow) user experience. We find users value a perception of responsiveness over stability.
Another disadvantage of repeated retranslation is
its computational inefficiency ($O(T^2)$)---one ends up
running very similar computations over and over again.
This is exacerbated with ASR intermediates that cause even more retranslations.

Simultaneous MT addresses both. The idea is that the model keeps a
state, and when a new transcription arrives from ASR, it processes it and then
decides whether it is ready to emit tokens
to the output or whether it requires more input.
This is for the case where emitted tokens are final and cannot be changed
anymore; we extend it towards emitting intermediates as well in \Cref{handling-intermediate-input}.

The deal-breaking drawback of repeated retranslation is that it
does not work with TTS output, as we cannot ``rewrite''
audio.\footnote{The time machine has not been invented yet.} User-perceived latency is directly determined by the modeling and decoding latency.

\textbf{In summary, compared to ``faking it'' via repeated retranslation, simultaneous MT solves two problems: when translating to display, simultaneous MT primarily reduces compute, while when translating to audio, it aims to
reduce latency from one sentence at a time to a few words.}

\hypertarget{model-architecture}{%
\subsection{Recap: Streaming Model Architectures}\label{model-architecture}}

An MT model is composed of two parts: encoder and decoder. We used the standard Transformer \cite{VaswaniSPUJGKP17} architecture for both. The simultaneous-MT model, in additional to the probability
distribution over tokens, also a returns a READ/WRITE probability at each
timestamp: When the probability for WRITE is below a threshold, the model
has not seen sufficient input to generate the next token. In this case,
we will not generate a token to output, but instead ingest another input
token (in a real-time system, this might block until the next ASR event is received).

Several model architectures and training methods for the READ/WRITE classifier have been proposed.
The current state of the art approach is
Monotonic Infinite Look-back (MILK) \cite{milk}.
A deeper discussion about WRITE probability
can be found in \Cref{scoring-beams}.

\hypertarget{non-streaming-beam-search-baseline}{%
\subsection{Recap: Non-streaming Beam Search}\label{non-streaming-beam-search-baseline}}

Non-streaming beam search is well-known, and relatively
straight-forward. The ``beam'' is the set of the
best-scoring $k$ hypotheses during the incremental decoding process. A
non-streaming beam decoder first executes the encoder for an entire input
sequence. Then it repeatedly runs one step of the decoder for each of
the (up to) $k$ hypotheses in the current beam and expands each
by all $n$ words in the vocabulary. This yields up to $n \cdot k$
expanded hypotheses from which we select the top $k$ based on their aggregated score. While iterating through the top $k$ hypotheses, we set aside those that ended with an \verb+<EOS>+ or exceed a maximum length. Decoding terminates once $k$ hypotheses have been finished.

\hypertarget{streaming-mt-beam-search}{%
\section{Beam Search for Simultaneous MT}\label{streaming-mt-beam-search}}

Simultaneous MT does not have access to the full input sentence, because it has not been fully recognizer or even spoken yet.
Instead, it must incrementally encode the input as it arrives, and translate as much as it deems safe as determined by the READ/WRITE classifier (while considering intermediate ASR events and potential incomplete words).

This poses a conflict with beam search, because \textbf{each hypothesis in the
beam could have read a different number of tokens and it could have also
written a different number of tokens}. We handle this
by processing hypotheses in groups that have seen the same number of tokens
from the input. For each group, we continue expanding in a loop hypotheses classified to WRITE until only hypotheses remain that we want to
READ.
During the expansion step, we score the hypotheses
and always select the top $k$. Once all top $k$ hypotheses want to READ, we
perform that READ simultaneously for all (i.e., await the next ASR event). Once we read all input tokens (reached a sentence boundary), we perform a final WRITE expansion loop until $k$ hypotheses have generated \verb+<EOS>+ token or the maximum output
length was reached. Again, at any point in time,
all hypotheses in the beam have READ the same input tokens.

Readers familiar with a speech-recognition model called Recurrent Neural Transducer (RNN-T) \cite{rnnt} will recognize the similarity. The RNN-T's BLANK symbol corresponds to the READ class, and our decoder is similar to the decoder laid out in \cite{rnnt}.






\hypertarget{state-management}{%
\subsection{State Management}\label{state-management}}

Since ASR inputs arrive in real time, a ``read'' decision may refer to
an input token that has not been spoken or recognized yet, and therefore
require the decoder to yield the CPU until the next ASR event. This
means that the internal decoding state must be carried across
incremental invocations of the decoder.

Specifically, we need to maintain the list of the current hypotheses,
and for each hypothesis, we need to store the encoder and decoder model
state, and couple other variables. We will refer to this as the beam
search state. The model state is necessary to store, so that we don't
need to recompute it every time a new token comes in. The encoder and
decoder model state caching is implemented using KV-cache mechanism \cite{kvcache}
which caches keys and values inside the transformer layer
computed in the previous timestamps.

\hypertarget{emitting-output}{%
\subsection{Emitting Output}\label{emitting-output}}

We want to write to the display intermediate outputs while we are
decoding input to improve user-perceived latency. We have two types of
outputs---intermediate and final. A final output is a prefix of the
output that will not change---we won't rewrite it. Intermediate output
can change, so the user might experience a bit of a flicker when words
get overwritten on the display.

To determine \textbf{final events}, at each decoding step, we take a
look at the top k hypotheses, and determine the common prefix of those.
That common prefix is then emitted as a final event since this prefix
can no longer change during the following steps of the beam search.

To generate \textbf{intermediate events}, each time ASR sends us an
event, we process it until all hypotheses end in a ``read'', then we
take the top hypothesis, and return that as an intermediate input. While
the common prefix of the hypotheses is stable---it won't change---the
rest of the output sequence in the top hypothesis can change, so we
treat it as intermediate.


\hypertarget{handling-intermediate-input}{%
\subsection{Handling Intermediate
Input}\label{handling-intermediate-input}}

The ASR engine itself is also running beam search similar to ours, and
it is therefore generating intermediate results which are not stable and
might change. To handle intermediate ASR inputs, we checkpoint the beam
search state for the last finalized input, run decoding for the
intermediate input, emit the top hypothesis as intermediate output, and
then rewind the state to the last finalized/stable state checkpoint.
This ensures that the user is getting more feedback about the
translation from the system, and we don't recompute the same state
multiple times.

\hypertarget{incomplete-words-from-asr}{%
\subsection{Incomplete Words from ASR}\label{incomplete-words-from-asr}}

An interesting edge case that we had not anticipated initially was that
ASR can return incomplete word prefixes in the stream of inputs. For
example, the word token \texttt{\_carpet} might be recognized as two word pieces \texttt{\_car} followed by \texttt{pet}, and these may be emitted in two
consecutive final events.
The issue
with incomplete words is that the tokenization of the word can change
once we get the rest of the word---while the ASR vocabulary may not
have a whole-word entry for \texttt{\_carpet}, the MT vocabulary may. The way
we handle this is that we take the input, and treat the prefix that ends
at a word boundary (space or punctuation) as finalized input. We cannot
know whether the remainder of the input string has stable tokenization,
so we treat it as intermediate input.

\hypertarget{scoring-beams}{%
\subsection{Scoring Beams}\label{scoring-beams}}

To rank beams, we need to compute the score for each beam. It is not
completely clear how the score should be computed in the case of
streaming NMT model. In case of RNN-T, the score of a beam is the score
of the path including the ``read'' or ``write'' probabilities. According
to pattern-recognition theory, error rate is minimized by selecting the
hypothesis with the highest probability of having given rise to the
observed input (Maximum-A-Posteriori, or MAP, decoding). This is
rigorously reflected by the RNN-T decoding algorithm, and also in the
RNN-T training loss. On the other hand, the ``write'' probability
provided by MILK \cite{milk} method used to train the streaming-MT model
is more of a heuristic. Up to now, MILK models have only been decoded
with greedy search, where an accurate probabilistic score for multiple
competing hypotheses is not needed.

We experimented with including the heuristic write probability in
the path score of a beam. This did not work well. The issue is that the
MILK loss does not include the write
probability\footnote{We hope that a future version of the MILK loss
model can provide a more rigorous ``write'' probability estimate that
enables the use of a true MAP decoder as shown to be optimal by
pattern-recognition theory.}, just the token probabilities (plus a latency penalty that
only indirectly affects the write probability). The approach that
we settled for was that the score of a beam is length normalized token
sequence probability, excluding the write probability at all.
Length normalization is needed because each beam could perform a
different number of writes.

\hypertarget{batching}{%
\subsection{Batching}\label{batching}}

Model inference is dominated by memory bandwidth more so than compute
cycles. Therefore, batching multiple model evaluations significantly
improves the inference time. We don't need to perform batching on the
encoder side because all of the beams read at the same time, which means
that we can hold the same encoder state for all of the beams. However, every
beam has a different decoder state because each beam could perform a
different number of writes.
In non-streaming beam search, batching these is easy
because all hypotheses share the same length. In the streaming case,
it's not so easy. When we are performing self-attention, the token
sequence is different for each beam which makes it hard to store all of
the beams in the same tensor.\footnote{This problem could be mitigated using LSTM
decoder, but that might introduce accuracy regression, so it is about a
trade-off.}

\hypertarget{handling-end-of-sentence-token-eos-on-the-inputoutput}{%
\subsection{Handling Sentence Boundaries}\label{handling-end-of-sentence-token-eos-on-the-inputoutput}}

The model sometimes produces \verb+<EOS>+ token before we get \verb+<EOS>+ token on input.
We decided not to allow the model to do that because we still want to
finish reading the whole sentence on the input.

Once we read \verb+<EOS>+ on input, we cannot perform any reads. We can only
write. In this case, we ignore the model's write
probability, and just keep writing until the model produces \verb+<EOS>+ token.
This strategy proved to be better than the strategy where we keep
writing until the write probability is above a threshold. Sometimes the
model does not produce \verb+<EOS>+ token at all. In that case, we want to stop
writing at some point. The heuristic that we settled for is that we have
a fixed threshold on the output sequence length that is based on the
number of input tokens. The threshold is equal to $ax + b$, where x
is the number of input tokens, and a and b coefficients that were
tweaked using cross-validation.

\hypertarget{experimental-results}{%
\section{Experimental Results}\label{experimental-results}}

This section presents results for various benchmarks that we performed
to evaluate the performance of the streaming beam search.
All of the benchmarks were performed on an embedded device
where we constrained it to use a single
$2.84$ GHz Qualcomm Snapdragon XR2 Kryo 585 CPU core.

\hypertarget{metrics}{%
\subsection{Metrics}\label{metrics}}

The main trade off that we are making when designing a streaming system
is between latency and accuracy. In MT, accuracy is commonly measured by the
\textbf{BLEU}\cite{bleu} score,
for ``bilingual evaluation understudy''. In our case, we only consider
final output tokens for BLEU score evaluation. We do not take into
account intermediate translations in BLEU score calculation.

We also want to measure the user's perceived latency.
Common metric used in simultaneous translation is
\textbf{word average lag}\cite{lag}
which estimates how many future input words on average we
need to read in order to translate an input word.

From the user perspective, it also matters how often we rewrite (change)
words on the display. For users, it is better when we don't rewrite any
words so they don't have to go back and reread the sentence. We came up
with a metric called \textbf{character flicker} which quantifies what
percentage of characters change on the display on average after each
write to the display.

The last metric that we used is \textbf{total runtime}. This measures
CPU time running the whole translation pipeline. The computational
resources are limited on device and every subsystem running on the
device has a certain compute budget, and when MT exceeds the budget, it
will get preempted meaning that we don't be allowed to run MT model for
a while on the device. Additionally, the total runtime also contributes
to the user's perceived latency. The total runtime also
includes the compute needed for processing intermediate inputs.

\hypertarget{benchmarks-without-intermediates}{%
\subsection{Benchmarks Without Intermediate Inputs}\label{benchmarks-without-intermediates}}

In these benchmarks we simulated streaming ASR input by splitting a
sentence into words, and then feeding one word after another to the
streaming system as an ASR final event. In this case, we only have final
inputs. We don't have intermediate inputs. We used internal
English-Spanish dataset for these benchmarks.

\hypertarget{bleu-comparison}{%
\subsubsection{Accuracy Comparison}\label{bleu-comparison}}

\Cref{t-bleu} shows BLEU score comparison between streaming and non-streaming beam search.
We can see around 10\% relative degradation between streaming
and non-streaming model which is expected because
non-streaming model sees the whole sentence before it starts to
translate. We observed that just by switching from non-causal to
causal transformer encoder, we lose around $2$ BLEU points.
We can also notice that streaming beam search improves the BLEU
score by $0.9$ compared to greedy decoding.

\begin{table}[ht!]
    \vspace{-2mm}
    \centering
    \footnotesize
    \setlength{\tabcolsep}{4pt}
    \begin{tabular}{ c | c | c }
        Beam & Streaming &  Non-streaming \\
        \hline
        1 & 45.7 & 52.5 \\
        \hline
        2 & 46.4 & 52.8 \\
        \hline
        3 & 46.6 & 52.9 \\
    \end{tabular}
    \caption{BLEU score comparison between a streaming and non-streaming MT model trained on the same data.}
    \label{t-bleu}
    \vspace{-10mm}
\end{table}

\hypertarget{runtime-comp}{%
\subsubsection{Runtime Comparison}\label{runtime-comp}}

\Cref{t-runtime} shows results for streaming and fake streaming by repeated retranslations.
We can observe that streaming performs much better in terms of character flicker.
Greedy streaming has no flicker because we don't have any intermediates.
It also has the lowest CPU time and average lag.
We can see that CPU time and average lag grows for streaming
when we increase the beam size.

Streaming beam search shows better results compared to fake streaming when we work with longer
sequences. Second part of \Cref{t-runtime} shows the comparison for sequences longer than ten words.
Character flicker is much better for streaming.
24\% character flicker basically means
that one quarter of the displayed words are overwritten. When it comes to CPU time, we can
see significant improvement for beam size two. The CPU time for beam size three is the same.

\begin{table}[ht!]
    \vspace{-2mm}
    \centering
    \footnotesize
    \setlength{\tabcolsep}{4pt}
    \begin{tabular}{ c | c | c | c | c | c | c }
        & \multicolumn{2}{c|}{CPU time (sec)} & \multicolumn{2}{c|}{Average lag} & \multicolumn{2}{c}{Char flicker (\%)} \\ \hline
        Beam & Streaming & Fake & Streaming & Fake & Streaming & Fake \\
        \hline
        1 & 3.0 & 4.2 & 4.8 & 6.1 & 0 & 18.9 \\
        \hline
        2 & 4.8 & 4.5 & 6.3 & 6.1 & 3.7 & 19.0 \\
        \hline
        3 & 6.6 & 4.7 & 6.8 & 6.1 & 3.7 & 19.7 \\
        \hline
        \multicolumn{7}{c}{Sequences longer than 10 words only} \\
        \hline
        2 & 13.5 & 18.2 & 8.7 & 13.0 & 2.5 & 24.0 \\
        \hline
        3 & 19.2 & 19.1 & 9.9 & 13.0 & 3.2 & 23.0 \\
    \end{tabular}
    \caption{Runtime comparison between streaming and fake streaming beam search with different beam sizes and no intermediate inputs.}
    \label{t-runtime}
    \vspace{-10mm}
\end{table}

\hypertarget{benchmarks-on-asr-data}{%
\subsection{Benchmarks on ASR data}\label{benchmarks-on-asr-data}}

In these benchmarks, we transcribed a set of audio recordings
from real-life English-Spanish conversations using ASR model and logged
the intermediate and final outputs from ASR. Then we fed both
intermediates and finals into the translation system in the same way the
model would receive them during real-time operation. We compared
streaming beam search and fake streaming with repeated retranslations.

\Cref{t-stream-asr} shows how long it takes to process a single input event
(intermediate or final), on average, from ASR, and character flicker on
the display.

\begin{table}[!ht]
    \vspace{-2mm}
    \centering
    \footnotesize
    \setlength{\tabcolsep}{4pt}
    \begin{tabular}{ c | c | c | c | c }
        & \multicolumn{2}{c | }{CPU time (ms)} & \multicolumn{2}{c}{Char flicker (\%)} \\ \hline
        Beam & Streaming & Fake & Streaming & Fake \\ \hline
        1 & 369 & 597 & 13.3 & 34.1 \\ \hline
        2 & 506 & 657 & 13.2 & 35.8 \\ \hline
        3 & 645 & 702 & 14.2 & 35.5 \\
    \end{tabular}
    \caption{Streaming beam search with different beam sizes including intermediate inputs.}
    \label{t-stream-asr}
    \vspace{-5mm}
\end{table}

We can see that CPU time is better for streaming compared to fake
streaming for all beam sizes. This is because the fake streaming needs
to perform more retranslations when we introduce intermediate input
events. The improvement is biggest for the beam size one. Flicker is
also much better for all beam sizes for streaming compared to fake
streaming.

\vspace{-2mm}
\section{Conclusions}

We have presented an adaptation of the beam-search algorithm, traditionally used in machine translation, to operate within a cascaded real-time speech translation system. 
We have discussed solutions to the following challenges in the process of adaptation by delving into (1) real-time processing of incomplete words from ASR, (2) minimizing user-perceived latency, (3) managing beam search hypotheses of unequal length and different model state, and (4) handling sentence boundaries. 

Our streaming beam decoder yielded significant improvements. Beam search increased the BLEU score by 1 point over greedy search, and compared to the simplistic heuristic of repeated input retranslation, it
reduced CPU time by up to 40\%, and decreased the character flicker rate by over 20\%.

We hope that our work provided a valuable contribution to simultaneous machine translation, offering a more efficient and effective approach.
Additionally, we hope that this paper can help guide the reader to navigate the problems that arise when implementing streaming MT beam search for a real-life system.

\clearpage

\bibliographystyle{IEEEtran}
\bibliography{mybib}

\end{document}